%% file: proximity_vi.tex
\pdfoutput=1

\documentclass{article}

\usepackage[final]{nips_2017}

\input{preamble}
\input{preamble_math}

\input{preamble_acronyms}

\title{Proximity Variational Inference}

\author{
  Jaan Altosaar\\
  Department of Physics\\
  Princeton University\\
  \texttt{altosaar@princeton.edu} \\
  \And
  Rajesh Ranganath\\
  Department of Computer Science\\
  Princeton University\\
  \texttt{rajeshr@cs.princeton.edu} \\
  \And
  David M.~Blei\\
  Data Science Institute\\
  Departments of Computer Science and Statistics\\
  Columbia University\\
  \texttt{david.blei@columbia.edu} \\
}

\begin{document}
\maketitle

\input{sec_abstract}
\input{sec_intro}
\input{sec_method}
\input{sec_experiments}
\input{sec_discussion}

\bibliographystyle{apalike}
\bibliography{bib}

\end{document}

%% file: preamble.tex
\usepackage[T1]{fontenc}
\usepackage{tgtermes}
\usepackage{amsmath} 
\usepackage{enumitem}

\usepackage{amssymb}
\newcommand{\mathbold}[1]{\ensuremath{\boldsymbol{\mathbf{#1}}}}

\usepackage{scalefnt,letltxmacro}
\LetLtxMacro{\oldtextsc}{\textsc}
\renewcommand{\textsc}[1]{\oldtextsc{\scalefont{1.10}#1}}
\usepackage[ttdefault=true]{AnonymousPro}

\usepackage[usenames,dvipsnames]{xcolor}
\definecolor{shadecolor}{gray}{0.9}

\usepackage{etoolbox,siunitx}
\robustify\bfseries
\usepackage[final,expansion=alltext]{microtype}
\usepackage[english]{babel}
\usepackage[parfill]{parskip}
\usepackage{afterpage}
\usepackage{framed}
\usepackage{nicefrac}

\usepackage{lineno}

\usepackage{ragged2e}

\DeclareRobustCommand{\parhead}[1]{\textbf{#1}~}

\usepackage{graphicx}
\usepackage[labelfont=bf]{caption}
\usepackage[format=hang]{subcaption}

\usepackage{booktabs}

\usepackage{natbib}

\usepackage[algoruled]{algorithm2e}
\usepackage{listings}
\usepackage{fancyvrb}
\fvset{fontsize=\normalsize}

\usepackage[colorlinks,linktoc=all]{hyperref}
\hypersetup{citecolor=Violet}
\hypersetup{linkcolor=black}
\hypersetup{urlcolor=MidnightBlue}

\usepackage{url}

\usepackage[nameinlink]{cleveref}

\usepackage[acronym,smallcaps,nowarn]{glossaries}
\glsdisablehyper{}

\usepackage{listings}
\lstdefinestyle{alp_style}{
    commentstyle=\color{OliveGreen},
    numberstyle=\tiny\color{black!60},
    stringstyle=\color{BrickRed},
    basicstyle=\ttfamily\scriptsize,
    breakatwhitespace=false,
    breaklines=true,
    captionpos=b,
    keepspaces=true,
    numbers=none,
    numbersep=5pt,
    showspaces=false,
    showstringspaces=false,
    showtabs=false,
    tabsize=2
}

%% file: preamble_math.tex
\DeclareRobustCommand{\mb}[1]{\mathbold{#1}}

\renewcommand{\mid}{~\vert~}

\newcommand{\mbx}{\mb{x}}

\newcommand{\mbz}{\mb{z}}

\newcommand{\mblambda}{\mb{\lambda}}

\newcommand{\E}{\mathbb{E}}

\newcommand{\cL}{\mathcal{L}}

%% file: preamble_acronyms.tex
\newacronym{KL}{kl}{Kullback-Leibler}
\newacronym{ELBO}{elbo}{evidence lower bound}
\newacronym{SVI}{svi}{stochastic variational inference}
\newacronym{GMM}{gmm}{Gaussian mixture model}
\newacronym{LDA}{lda}{latent Dirichlet allocation}
\newacronym{VI}{vi}{variational inference}
\newacronym{PVI}{pvi}{proximity variational inference}
\newacronym{BBVI}{bbvi}{black box variational inference}

%% file: sec_abstract.tex

\begin{abstract}

Variational inference is a powerful approach for approximate posterior
inference. However, it is sensitive to initialization and can be subject to poor
local optima. In this paper, we develop \gls{PVI}. \gls{PVI} is a new method for
optimizing the variational objective that constrains subsequent iterates of the
variational parameters to robustify the optimization path. Consequently,
\gls{PVI} is less sensitive to initialization and optimization quirks
and finds better local optima. We demonstrate our method on three proximity statistics.
We study \gls{PVI}
on a Bernoulli factor model and sigmoid belief network with both real and
synthetic data and compare to deterministic annealing~\citep{Katihara:2008}. We
highlight the flexibility of \gls{PVI} by designing a proximity
statistic for Bayesian deep learning models such as the variational
autoencoder~\citep{kingma2014autoencoding,rezende2014stochastic}. Empirically, we show that \gls{PVI}
consistently finds better local optima and gives better predictive performance.

\end{abstract}

%% file: sec_intro.tex

\section{Introduction} \label{sec:introduction}

\Gls{VI} is a powerful method for probabilistic modeling.  \gls{VI} uses
optimization to approximate difficult-to-compute conditional
distributions~\citep{jordan1999introduction}.  In its modern incarnation, it has
scaled Bayesian computation to large data sets~\citep{hoffman2013stochastic},
generalized to large classes of
models~\citep{kingma2014autoencoding,ranganath2014black,Rezende:2015}, and has been
deployed as a computational engine in probabilistic programming
systems~\citep{DBLP:journals/corr/MansinghkaSP14,Kucukelbir:2015,tran2016edward}.

Despite these significant advances, however, \gls{VI} has drawbacks.
For one, it tries to iteratively solve a difficult nonconvex
optimization problem and its objective contains many local optima.
Consequently, \gls{VI} is sensitive to initialization and easily
gets stuck in a poor solution.  We develop a new
optimization method for \gls{VI}.  Our method is less sensitive to
initialization and finds better optima.

Consider a probability model $p(\mbz,\mbx)$ and the goal of
calculating the posterior $p(\mbz \mid \mbx)$. The idea behind \gls{VI}
is to posit a family of distributions over the hidden variables
$q(\mbz ; \mblambda)$ and then fit the variational parameters $\mblambda$
to minimize the \gls{KL} divergence between the approximating family
and the exact posterior,
$\textsc{kl}(q(\mbz ; \mblambda) || p(\mbz \mid \mbx))$. The \gls{KL} is
not tractable so \gls{VI} optimizes a proxy.  That proxy is the
\gls{ELBO},
\begin{align}
\cL(\mblambda) = \E[\log p(\mbz,\mbx)] - \E[\log q(\mbz ; \mblambda)],
\label{eq:ELBO}
\end{align}
where expectations are taken
with respect to $q(\mbz ; \mblambda)$. Maximizing the \gls{ELBO} with respect to
$\mblambda$ is equivalent to minimizing the \gls{KL} divergence.

\input{fig_arrows}

The issues around \gls{VI} stem from the \gls{ELBO} and the iterative algorithms
used to optimize it.  When the algorithm zeroes (or nearly zeroes) some of the
support of $q(\mbz ; \mblambda)$, it becomes hard to later ``escape,'' i.e., to
add support for the configurations of the latent variables that have been
assigned zero probability~\citep{MacKay:2003, Burda2016}.  This leads to poor
local optima and to sensitivity to the starting point, where a misguided
initialization will lead to such optima.  These problems happen in both
gradient-based and coordinate ascent methods. We address these issues with
\glsreset{PVI} \gls{PVI}, a variational inference algorithm that is specifically designed to
avoid poor local optima and to be robust to different initializations.

\gls{PVI} builds on the proximity perspective of gradient ascent.  The proximity
perspective views each step of gradient ascent as a constrained minimization of a Taylor expansion
of the objective around the previous step's
parameter~\citep{Spall:2003,boyd2004convex}.  The constraint, a
\textit{proximity constraint}, enforces that the next point should be inside a
Euclidean ball of the previous.  The step size relates to the size of that
ball.

In \gls{VI}, a constraint on the Euclidean distance means that all
dimensions of the variational parameters are equally constrained.
We posit that this leads to problems; some dimensions need more
regularization than others. For example, consider a variational
distribution that is Gaussian.  A good optimization will change the
variance parameter more slowly than the mean parameter to prevent
rapid changes to the support.  The Euclidean constraint cannot enforce
this. Furthermore, the constraints enforced by gradient descent are transient; the constraints are relative to the previous iterate---one
poor move during the optimization can lead to permanent optimization problems.

To this end, \gls{PVI} uses proximity constraints that are more meaningful to
variational inference and to optimization of probability parameters.  A
constraint is defined using a proximity statistic and distance function. As one
example we consider a constraint based on the entropy proximity statistic. This
limits the change in entropy of the variational approximation from one step to
the next.  Consider again a Gaussian approximation. The entropy is a function of
the variance alone and thus the entropy constraint counters the pathologies
induced by the Euclidean proximity constraint. We also study constraints built from other proximity statistics, such as those that penalize the rapid changes in the mean and variance of the approximate posterior.

\Cref{fig:bernoulli_arrows} provides an illustration of the advantages of
\gls{PVI}.  Our goal is to estimate the parameters of a factor analysis model
with variational inference, i.e., using the posterior expectation under a fitted
variational distribution.  We run variational inference $100$ times, each time
initializing the estimates (the model parameters) to a different position
on a ring around the truth.

In the figure, red points indicate the true value. The start locations of the
green arrows indicate the initialized estimates. Green points indicate the final
estimates, after optimizing from the initial points. Panel (a) shows that
optimizing the standard \gls{ELBO} with gradients leads to poor local optima and
misplaced estimates.  Panel (b) illustrates that regardless of the initialization, \gls{PVI} with an entropy proximity statistic find estimates that are close to the true value.

The rest of the paper is organized as follows. \Cref{sec:variational_inference}
reviews variational inference and the proximity perspective of gradient
optimization.  \Cref{sec:pvi} derives \gls{PVI}; we develop four proximity
constraints and two algorithms for optimizing the \gls{ELBO}.
We study three models in \Cref{sec:experiments}: a Bernoulli factor
model, a sigmoid belief network \citep{Mnih:2016:VIM:3045390.3045621}, and a
variational autoencoder~\citep{kingma2014autoencoding,rezende2014stochastic}.  On the MNIST data set, \gls{PVI} generally outperforms classical methods for variational inference.
\input{sec_related}
\section{Variational inference}
\label{sec:variational_inference}

Consider a model $p(\mbx, \mbz)$, where $\mbx$ is the observed data
and $\mbz$ are the latent variables.  As described in
\Cref{sec:introduction}, \gls{VI} posits an approximating family
$q(\mbz; \mblambda)$ and maximizes the \gls{ELBO} in \Cref{eq:ELBO}.
Solving this optimization is equivalent to finding the variational
approximation that minimizes \gls{KL} divergence to the exact
posterior \citep{jordan1999introduction,wainwright2008graphical}.

\subsection{Gradient ascent has Euclidean proximity}
\label{sec:euclidean_proximity}
Gradient ascent maximizes the \gls{ELBO} by repeatedly following its
gradient.  One view of this algorithm is that it repeatedly maximizes
the linearized \gls{ELBO} subject to a proximity constraint on the
current variational parameter~\citep{Spall:2003}. The name `proximity' comes from constraining subsequent parameters to remain
close in the proximity statistic. In gradient ascent, the
proximity statistic for the variational parameters is the identity function
$f(\mblambda) = \mblambda $, and the distance function is the square
difference.

Let $\mblambda_t$ be the variational parameters at iteration $t$ and
$\rho$ be a constant. To obtain the next iterate $\mblambda_{t+1}$,
gradient ascent maximizes the linearized \gls{ELBO},
\begin{align}
  \label{eq:linearized-elbo}
  \begin{split}
    U(\mblambda_{t+1}) =
    &\cL(\mblambda_t) +
    \nabla\cL(\mblambda_t)^\top(\mblambda_{t+1}-\mblambda_t) -
    \frac{1}{2\rho} (\mblambda_{t+1}-\mblambda_t)^\top
    (\mblambda_{t+1}-\mblambda_t).
  \end{split}
\end{align}
Specifically, this is the linearized \gls{ELBO} around $\mblambda_t$
subject to $\mblambda_{t+1}$ being close in squared Euclidean distance
to $\mblambda_t$.

Finding the $\mblambda_{t+1}$ which maximizes
\Cref{eq:linearized-elbo} yields
\begin{align}
  \mblambda_{t+1} = \mblambda_t + \rho
  \nabla\cL(\mblambda_t). \label{eq:standard_update}
\end{align}
This is the familiar gradient ascent update with a step size of $\rho$. The step size $\rho$ controls the radius of
the Euclidean ball which demarcates valid next steps for the
parameters.  Note that the Euclidean constraint between subsequent
iterates is implicit in all gradient ascent algorithms.

\subsection{An example where variational inference fails}
\label{sec:bernoulli_factor_model}

We study a setting where variational inference suffers from
poor local optima. Consider a factor model, with latent variables $z_{ik}~\sim~\textrm{Bernoulli}(\pi)$ and data $x_i~\sim~\textrm{Gaussian}\left(\mu = \textstyle \sum_k z_{ik}
        \mu_k, \sigma^2=1\right)$. This is a ``feature'' model of real-valued data $x$; when one of the
features is on (i.e., $z_{ik}=1$), the $i$th mean shifts according the
that feature's mean parameter (i.e., $\mu_{k}$).  Thus the binary
latent variables $z_{ik}$ control which cluster means $\mu_k$
contribute to the distribution of $x_i$.

The Bernoulli prior is parametrized by $\pi$; we choose a Bernoulli
approximate posterior
$q(z_k;\lambda_k)~=~\textrm{Bernoulli}(\lambda_{k})$.  A common
approach to \gls{VI} is coordinate ascent~\citep{Bishop:2006}, where we
iteratively optimize each variational parameter.  The optimal
variational parameter for $z_{ik}$ is
\begin{align}
  \label{eq:z_update} \lambda_{ik} \propto
   \exp \left\{\E_{-z_{ik}} \left[- \frac{1}{2\sigma^2}(x_i -
    \sum_{j} z_{ij} \mu_j)^2\right]\right\}.
\end{align}
We can use this update in a variational EM setting.  The corresponding
gradient for $\mu_k$ is
\begin{align}
  \label{eq:mu-gradient}
  \frac{\partial \cL}{\partial\mu_k}
  &= -
    \frac{1}{\sigma^2}\sum_i
    \left(-x_i \lambda_{ik} + \lambda_{ik} \mu_k
    + \lambda_{ik} \sum_{j \neq k}\lambda_{ij}\mu_j\right).
\end{align}

Meditating on these two equations reveals a deficiency in mean-field
variational inference.  First, if the mean parameters $\mu$ are
initialized far from the data then $q^*(z_{ik} = 1)$ will be very
small.  The reason is in \Cref{eq:z_update}, where the squared
difference between the data $x_i$ and the expected cluster mean will
be large and negative.  Second, when the probability of cluster
assignment is close to zero, $\lambda_{ik}$ is small.  This means that
the norm of the gradient in \Cref{eq:mu-gradient} will be small.
Consequently, learning will be slow. We see this phenomenon in \Cref{fig:bernoulli_arrows} (a). Variational inference
arrives at poor local optima and does not always recover the correct cluster
means $\mu_k$.

%% file: fig_arrows.tex

\begin{figure*}[htb!]
\centering
\begin{subfigure}{0.45\linewidth}
  \centering
  \includegraphics[width=\linewidth]{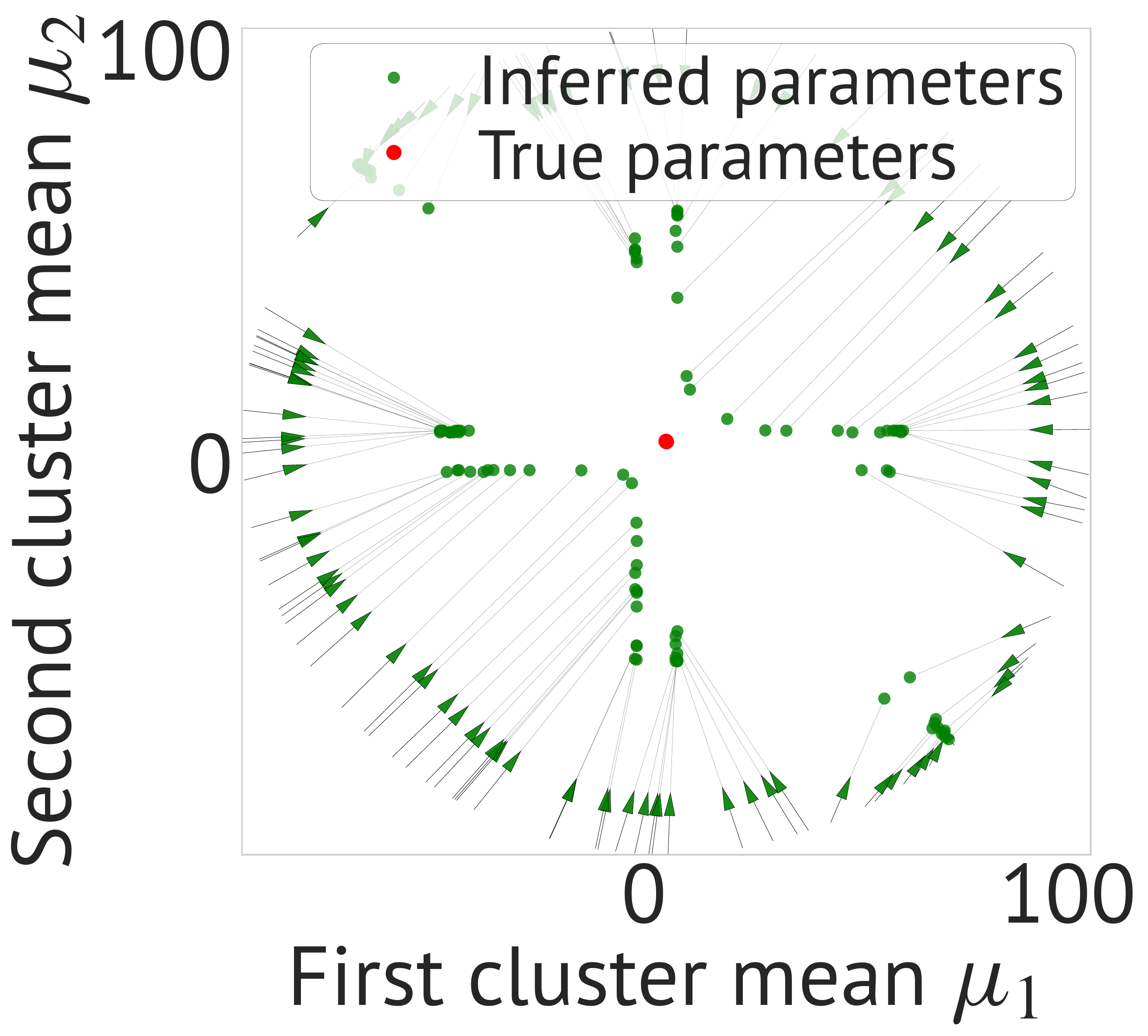}
  \label{fig:bernoulli_vanilla}
  \caption{Variational inference}
\end{subfigure}
\begin{subfigure}{0.45\linewidth}
    \centering
    \includegraphics[width=\linewidth]{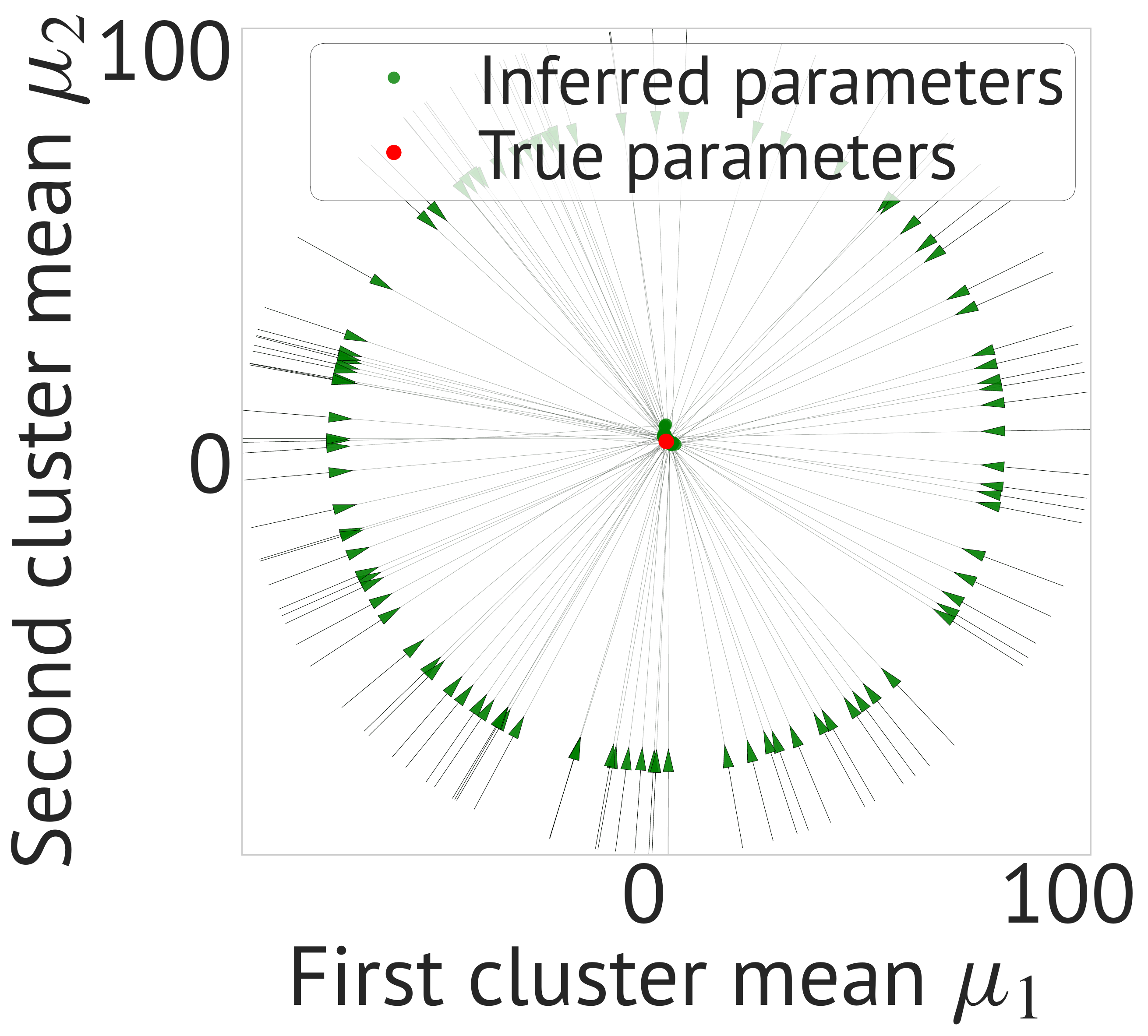}
    \label{fig:bernoulli_global}
    \caption{Proximity variational inference, \Cref{algo:global}}
\end{subfigure}
\caption{\textbf{\glsreset{PVI}\Gls{PVI} is robust to bad
    initialization.} We study a Bernoulli factor model. Model parameters are randomly
  initialized on a ring around the known true parameters (in red) used
  to generate the data. The arrows start at these parameter
  initializations and end at the final parameter estimates (shown as
  green dots). \textbf{(a)} Variational inference with gradient ascent suffers
  from multiple local optima and cannot reliably recover the
  truth. \textbf{(b)} \gls{PVI} with an entropy proximity statistic reliably infers the
  true parameters using \Cref{algo:global}.}
\label{fig:bernoulli_arrows}
\end{figure*}

%% file: sec_related.tex

\parhead{Related work.}  Recent work has proposed several related algorithms. \citet{khan2015kullback} and \citet{Theis2015} develop a method to optimize the \gls{ELBO} that imposes a soft limit on the change in \gls{KL} of consecutive variational approximations. This is equivalent to \gls{PVI} with identity proximity statistics and a \gls{KL} distance function. \citet{Khan:2016:FSV:3020948.3020982} extend both prior works to other
divergence functions. Their general approach is equivalent to \gls{PVI} identity proximity statistics and distance functions given by strongly-convex divergences. Compared to prior work, \gls{PVI} generalizes to a broader class of
proximity statistics. We develop proximity statistics based on entropy,
\gls{KL}, orthogonal weight matrices, and the mean and variance of the
variational approximation.

The problem of model pruning in variational inference has
also been studied and
analytically solved in a matrix factorization model in \citet{nakajima2013}---this method is model-specific, whereas \gls{PVI} applies to a much broader class of latent variable models. Finally, deterministic annealing~\citep{Katihara:2008} consists of adding a temperature parameter to the entropy term in the \gls{ELBO} that is annealed to one during inference. This is similar to \gls{PVI} with the entropy proximity statistic which keeps the entropy stable across iterations. Deterministic annealing enforces global penalization of low-entropy configurations of latent variables rather than the smooth constraint used in \gls{PVI}, and cannot accommodate the range of proximity statistics we design in this work.

%% file: sec_method.tex
\section{Proximity variational inference}
\label{sec:pvi}

\glsreset{PVI}

\input{algo_local}

We now develop \gls{PVI}, a variational inference method that is robust to
initialization and can consistently reach good local optima
(\Cref{sec:method}).  \gls{PVI} alters the notion of proximity. We further
restrict the iterates of the variational parameters by deforming the Euclidean
ball implicit in classical gradient ascent. This is done by choosing proximity
statistics that are not the identity function, and distance functions that are
different than the square difference. These design choices help guide the
variational parameters away from poor local optima
(\Cref{sec:proximity_examples}).  One drawback of the proximity perspective is
that it requires an inner optimization at each step of the outer optimization.
We use a Taylor expansion to avoid this computational burden
(\Cref{sec:proximity_taylor}).

\subsection{Proximity constraints for variational inference}
\label{sec:method}
\gls{PVI} enriches the proximity constraint in gradient ascent of the
\gls{ELBO}.  We want to develop constraints on the iterates $\mblambda_t$
to counter the pathologies of standard variational inference.

Let $f(\cdot)$ be a \emph{proximity statistic}, and let $d$ be a differentiable
distance function that measures distance between proximity statistic iterates. A
\emph{proximity constraint} is the combination of a distance function $d$ applied
to a proximity statistic $f$. (Recall that in classical gradient ascent,
the Euclidean proximity constraint uses the identity as the proximity statistic
and the square difference as the distance.)
Let $k$ be the scalar magnitude of the proximity constraint. We
define the proximity update equation for the variational parameters
$\mblambda_{t+1}$ to be
\begin{align}
  \begin{split}
    U(\mblambda_{t+1}) = &\cL(\mblambda_t)
    + \nabla \cL(\mblambda_t)^\top (\mblambda_{t+1}-\mblambda_t)
    \\
    &-
    \frac{1}{2\rho}
    (\mblambda_{t+1}-\mblambda_t)^\top (\mblambda_{t+1}-\mblambda_t)
    - k \cdot d(f(\tilde\mblambda), f(\mblambda_{t+1})),
  \end{split}
  \label{eq:f-constrained}
\end{align}
where $\tilde\mblambda$ is the variational parameter to which we are measuring closeness. In gradient ascent, this is the previous parameter
$\tilde\mblambda = \mblambda_t$, but our construction can enforce proximity
to more than just the previous parameters. For example, we can set
$\tilde\mblambda$ to
be an exponential moving average\footnote{The exponential moving average of a variable $\mblambda$ is denoted $\tilde\mblambda$ and is updated according to $\tilde\mblambda \leftarrow \alpha\tilde\mblambda + (1-\alpha)\mblambda$, where $\alpha$ is a decay close to one.}---this adds robustness to one-update
optimization missteps.

The next parameters are found by maximizing \Cref{eq:f-constrained}.  This
enforces that the variational parameters between updates will remain
close in the proximity statistic $f(\mblambda)$.  For example, $f(\mblambda)$
might be the entropy of the variational approximation; this can avoid
zeroing out some of its support.  This procedure is detailed in
\Cref{algo:local}. The magnitude $k$ of the constraint is a
hyperparameter. The inner optimization loop optimizes the update equation $U$ at each step.

\input{algo_global}

\subsection{Proximity statistics for variational inference}
\label{sec:proximity_examples}
We describe four proximity statistics $f(\mblambda)$ appropriate for
variational inference.  Together with a distance function,
these proximity statistics yield proximity constraints.
(We study them in \Cref{sec:experiments}.)

\parhead{Entropy proximity statistic.} Consider a constraint built from the
entropy proximity statistic, $f(\mblambda) = \mathrm{H}(q(\mbz; \mblambda))$. Informally,
the entropy measures the amount of randomness present in a
distribution.  High entropy distributions look more uniform across
their support; low entropy distributions are peaky.

Using the entropy in \Cref{eq:f-constrained} constrains all updates to
have entropy close to their previous update. When the variational
distributions are initialized with large entropy, this statistic
balances the ``zero-forcing'' issue that is intrinsic to variational
inference~\citep{MacKay:2003}.  \Cref{fig:bernoulli_arrows} demonstrates how \gls{PVI} with an
entropy constraint can correct this pathology.

\parhead{\gls{KL} proximity statistic.} We can rewrite the \gls{ELBO} to
include the \gls{KL} between the approximate posterior and the prior
\citep{kingma2014autoencoding},
\begin{align*}
  \cL(\mblambda) =
  \E [ \log p(\mb{x} \mid \mb{z})] - \mathrm{KL}(q(\mbz \mid \mbx; \mblambda) || p(\mbz) ).
\end{align*}
Flexible models tend to minimize the \gls{KL} divergence too quickly and get
stuck in poor optima~\citep{DBLP:conf/conll/BowmanVVDJB16,2016arXiv160605579H}. The choice of \gls{KL} as a
proximity statistic prevents the \gls{KL} from being optimized
too quickly relative to the likelihood.

\parhead{Mean/variance proximity statistic.} A common theme in the problems
with variational inference is that the bulk of the probability mass can
quickly move to a point where that dimension will no longer be
explored~\citep{Burda2016}. One way to address this is to restrict the
mean and variance of the variational approximation to change slowly
during optimization.
This constraint only allows higher order moments of the variational
approximation to change rapidly. The mean $\mu=\E_{q(\mbz; \mblambda)}[\mbz]$ and variance
$\textrm{Var}(\mbz) = \E_{q(\mbz; \mblambda)}[(\mbz - \mu)^2]$ are the statistics
$f(\mblambda)$ we constrain.

\parhead{Orthogonal proximity statistic.} In Bayesian deep learning models such as the variational autoencoder~\citep{kingma2014autoencoding,rezende2014stochastic} it is common to parametrize the variational distribution with a neural network. Orthogonal weight matrices make optimization easier in neural networks by allowing gradients to propagate further~\citep{saxe2013exact}. We can exploit this fact to design an orthogonal proximity statistic for the weight matrices $W$ of neural networks: $f(W) = WW^\top$. With an orthogonal initialization for the weights, this statistic enables efficient optimization.

We gave four examples of proximity statistics that, together with a distance function, yield proximity constraints. We emphasize that any function of the variational parameters $f(\mblambda)$ can
be designed to ameliorate issues with variational inference.

\subsection{Linearizing the proximity constraint for fast proximity variational inference}
\label{sec:proximity_taylor}

\gls{PVI} in \Cref{algo:local} requires optimizing the update equation,
\Cref{eq:f-constrained}, at each
iteration. This rarely has a closed-form solution and requires a
separate optimization procedure that is computationally expensive.

\input{tab_sbn}

An alternative is to use a first-order Taylor expansion of the
proximity constraint. Let $\nabla d$ be the gradient with respect to
the second argument of the distance function, and $f(\tilde \mblambda)$
be the first
argument to the distance. We compute the expansion around
$\mblambda_t$ (the variational parameters at step $t$),
\begin{align*}
  U(\mblambda_{t+1}) = &\cL(\mblambda_t) + \nabla \cL(\mblambda_t)^\top(\mblambda_{t+1}-\mblambda_t)
                       - \frac{1}{2\rho} (\mblambda_{t+1}-\mblambda_t)^\top (\mblambda_{t+1}-\mblambda_t)\\
                       & - k \cdot (d(f(\tilde \mblambda), f(\mblambda_{t})) + \nabla d (f(\tilde \mblambda), f(\mblambda_t)) \nabla f(\mblambda_t)^\top (\mblambda_{t+1}-\mblambda_t)).
\end{align*}
This linearization enjoys a closed-form solution for
$\mblambda_{t+1}$,
\begin{align}
  \mblambda_{t+1} = \mblambda_t + \rho(\nabla \cL(\mblambda_t) - k \cdot (\nabla d (f(\tilde \mblambda), f(\mblambda_t)) \nabla f(\mblambda_t)).
  \label{eq:fast-update}
\end{align}
Note that setting $\tilde \mblambda$ to the current parameter $\mblambda_t$
removes the proximity constraint. Distance functions are minimized at
zero so their derivative is zero at that point.

Fast \gls{PVI} is detailed in \Cref{algo:global}. Unlike \gls{PVI} in \Cref{algo:local}, the
update in \Cref{eq:fast-update} does not require an inner
optimization. Fast \gls{PVI} is tested in \Cref{sec:case-study}.
The complexity of fast \gls{PVI} is similar to standard \gls{VI}
because fast \gls{PVI} optimizes the \gls{ELBO} subject to the
distance constraint in $f$. (The added complexity comes from computing the
derivative of $f$; no inner
optimization loop is required.)

Finally, note that fast \gls{PVI} implies a global objective which
varies over time.  It is
\begin{align*}
\begin{split}
  \cL_{\textrm{proximity}}(\mblambda_{t+1}) = &\E_q [ \log p(\mbx, \mbz)] - \E_q [ \log q(\mblambda_{t+ 1})] - k \cdot d(f(\tilde\mblambda), f(\mblambda_{t+1})).
\end{split}
\end{align*}
Because $d$ is a distance, this remains a lower bound on the evidence,
but where new variational approximations remain close in $f$ to
previous iterations' distributions.

%% file: algo_local.tex

\begin{algorithm}[t]
\DontPrintSemicolon
\KwIn{Initial parameters $\mblambda_0$, proximity statistic $f(\mblambda)$, distance function $d$}
\KwOut{Parameters $\mblambda$ of variational $q(\mblambda)$ that maximize the \gls{ELBO} objective}
\While{$\cL$ not converged} {
  $\mblambda_{t+1} \gets \mblambda_t + \textrm{Noise}$\\
  \While{U not converged} {
    Update $\mblambda_{t+1} \gets \mblambda_{t+1} + \rho \nabla_{\mblambda}U(\mblambda_{t+1})$
  }
  $\mblambda_t \gets \mblambda_{t+1}$\;
}
\Return{$\mblambda$}\;
\caption{\textbf{Proximity variational inference}}
\label{algo:local}
\end{algorithm}

%% file: algo_global.tex

\begin{algorithm}[t]
\DontPrintSemicolon
\KwIn{Initial parameters $\mblambda_0$, adaptive learning rate optimizer, proximity statistic $f(\mblambda)$, distance $d$}
\KwOut{Parameters $\mblambda$ of the variational distribution $q(\mblambda)$ that maximize the \gls{ELBO} objective}
\While{$\cL_\textrm{proximity}$ not converged} {
	$\mblambda_{t+1} = \mblambda_t + \rho(\nabla \cL(\mblambda_t) - k \cdot (\nabla d (f(\tilde \mblambda), f(\mblambda_t)) \nabla f(\mblambda_t)).$ \\
	$\tilde \mblambda = \alpha \tilde \mblambda + (1 - \alpha) \mblambda_{t+1}$
}
\Return{$\mblambda$}\;
\caption{\textbf{Fast proximity variational inference}}
\label{algo:global}
\end{algorithm}

%% file: tab_sbn.tex

\begin{table}[tb]
\centering
\label{table:sbn_1_layer}
\begin{tabular}{lSSSS}
\toprule
&\multicolumn{2}{c}{Bad initialization} & \multicolumn{2}{c}{Good initialization} \\
Inference method & \multicolumn{1}{c}{\gls{ELBO}} & \multicolumn{1}{c}{Marginal likelihood} & \multicolumn{1}{c}{\gls{ELBO}} & \multicolumn{1}{c}{Marginal likelihood} \\
\midrule
Variational inference & -663.7 & -636.7 & -122.1 & -114.1 \\
Deterministic annealing & -119.4 & -110.2 & -116.5 & -108.7 \\
\gls{PVI}, Entropy constraint & \bfseries -118.0 & \bfseries -110.0 & \bfseries -114.1 & \bfseries -107.5 \\
\gls{PVI}, Mean/variance constraint & -119.9 & -111.1 & -115.7 & -108.3 \\
\bottomrule
\end{tabular}
\vspace{1ex}
\caption{\textbf{\Acrlong{PVI} improves on deterministic annealing~\citep{Katihara:2008} and \gls{VI} in a one-layer sigmoid belief network.}
We report validation \glsreset{ELBO}\gls{ELBO} and marginal likelihood on the binary MNIST dataset~\citep{pmlr-v15-larochelle11a}. The model has one stochastic layer of $200$~latent
variables. \gls{PVI} outperforms the classical variational inference algorithm
and deterministic annealing~\citep{Katihara:2008}.}
\end{table}

%% file: sec_experiments.tex

\section{Case studies}
\label{sec:case-study}
\label{sec:experiments}

We developed \acrfull{PVI}. We now empirically study \gls{PVI},\footnote{Source code is available at \url{https://github.com/altosaar/proximity_vi}.}
variational inference, and deterministic
annealing~\citep{Katihara:2008}. To assess the robustness of
inference methods, we study both good and bad initializations.  We
evaluate whether the methods can recover from poor initialization, and
whether they improve on \gls{VI} in well-initialized models.

We first study sigmoid belief networks. We found \gls{VI} fails to
recover good solutions while \gls{PVI} and deterministic annealing
recover from the initialization, and improve over \gls{VI}. \gls{PVI} yields further
improvements over deterministic annealing in terms of held-out values
of the \gls{ELBO} and marginal likelihood. We then studied a
variational autoencoder model of images.  Using an orthogonal
proximity statistic, \gls{PVI} improves over classical
\gls{VI}.\footnote{We also compared \gls{PVI} to
  \citet{khan2015kullback}. Specifically, we tested \gls{PVI} on the
  Bayesian logistic regression model from that paper and with the same
  data.  Because Bayesian logistic regression has a single mode, all
  methods performed equally well.  We note that we could not apply
  their algorithm to the sigmoid belief network because it would
  require approximating difficult iterated expectations.}

\input{tab_sbn_3}

\paragraph{Hyperparameters.} For \gls{PVI}, we use the inverse Huber distance
for $d$.\footnote{We define the inverse Huber distance   $d(x, y)$ to be $|x -
y|$ if $|x - y| < 1$ and $0.5(x-y)^2 + 0.5$   otherwise. The constants ensure
the function and its derivative are   continuous at $|x-y| = 1$.} The inverse
Huber distance penalizes smaller values than the square difference. For
\gls{PVI} \Cref{algo:global}, we set the exponential moving average decay constant for $\tilde\mblambda$ to $\alpha=0.9999$. We set the constraint magnitude
$k$ (or temperature parameter in deterministic annealing) to the initial
absolute value of the \gls{ELBO} unless otherwise specified. We explore two
annealing schedules for \gls{PVI} and deterministic annealing: a linear decay
and an exponential decay. For the exponential decay, the value of the magnitude
at iteration $t$ of $T$ total iterations is set to $k\cdot \gamma^{\frac{t}{T}}$
where $\gamma$ is the decay rate. We use the Adam
optimizer~\citep{DBLP:journals/corr/KingmaB14}.

\subsection{Sigmoid belief network}

The sigmoid belief network is a discrete latent variable model with
layers of Bernoulli latent
variables~\citep{neal1992connectionist,ranganath2015deep}. It is
commonly used to benchmark variational inference
algorithms~\citep{Mnih:2016:VIM:3045390.3045621}. The approximate
posterior is a collection of Bernoullis, parameterized by an inference
network with weights and biases.  We fit these variational parameters
with \gls{VI}, deterministic annealing~\citep{Katihara:2008}, or
\gls{PVI}.

A pervasive problem in applied \gls{VI} is how to initialize parameters; thus to
assess the robustness of our method, we study good and bad initialization. For
good initialization, we set the Bernoulli prior to $\pi=0.5$ and use the
normalized initialization in \citet{DBLP:journals/jmlr/GlorotB10} for the
weights of the generative neural network. For bad initialization, the Bernoulli
prior is set to $\pi=0.001$ and the weights of the generative neural network are
initialized to~$-100$.

We learn the weights and biases of the model with gradient ascent. We
use a step size of $\rho=10^{-3}$ and train for $4\times10^6$
iterations with a batch size of $20$.  For \gls{PVI} \Cref{algo:global} and
deterministic annealing, we grid search over exponential decays with
rates
$\gamma\in \{10^{-5}, 10^{-6}, ..., 10^{-10}, 10^{-20}, 10^{-30}\}$
and report the best results for each algorithm.  (We also explored
linear decays but they did not perform as well.)  To reduce the
variance of the gradients, we use the leave-one-out control variate
of~\citet{Mnih:2016:VIM:3045390.3045621} with $5$ samples. (This is an
extension to black-box variational
inference~\citep{ranganath2014black}.)

\input{tab_dlgm}

\paragraph{Results on MNIST.} We train a sigmoid belief network model
on the binary MNIST dataset of handwritten
digits~\citep{pmlr-v15-larochelle11a}. For evaluation, we compute the
\gls{ELBO} and held-out marginal likelihood with importance sampling
($5000$ samples, as in \cite{rezende2014stochastic}) on the
validation set of $10^4$ digits. In \Cref{table:sbn_1_layer} we show
the results for a model with one layer of $200$ latent
variables. \gls{PVI} and deterministic annealing yield improvements in
the held-out marginal likelihood in comparison to \gls{VI}.
\Cref{table:sbn_3_layer} displays similar results for a three-layer
model with $200$~latent variables per layer. In both one and
three-layer models the \gls{KL} proximity statistic performs worse
than the mean/variance and entropy statistics; it requires different
decay schedules. Overall, \gls{PVI} with the entropy and mean/variance
proximity statistics yields better performance than \gls{VI} and
deterministic annealing.

\subsection{Variational autoencoder}
\label{sec:variational_autoencoder}

We study the variational autoencoder~\citep{kingma2014autoencoding,rezende2014stochastic} on binary MNIST data~\citep{pmlr-v15-larochelle11a}. The model has one layer of $100$ Gaussian latent variables. The inference network and generative network are chosen to have two hidden layers of size $200$ with rectified linear unit activation functions. We use an orthogonal initialization for the inference network weights. The learning rate is set to $10^{-3}$ and we run \gls{VI} and \gls{PVI} for $5\times 10^4$ iterations. The orthogonal proximity statistic changes rapidly during optimization, so we use constraint magnitudes $k \in \{1, 10^{-1}, 10^{-2}, ..., 10^{-5}\}$, with no decay, and report the best result. We compute the \gls{ELBO} and importance-sampled marginal likelihood estimates on the validation set. In \Cref{table:dlgm2} we show that \gls{PVI} with the orthogonal proximity statistic on the weights of the inference network enables easier optimization and improves over \gls{VI}.

%% file: tab_sbn_3.tex

\begin{table}[tb]
\centering
\begin{tabular}{lSSSS}
\toprule
&\multicolumn{2}{c}{Bad initialization} & \multicolumn{2}{c}{Good initialization} \\
Inference method & \multicolumn{1}{c}{\gls{ELBO}} & \multicolumn{1}{c}{Marginal likelihood} & \multicolumn{1}{c}{\gls{ELBO}} & \multicolumn{1}{c}{Marginal likelihood} \\
\midrule
Variational inference & -504.5 & -464.3 & -117.4 & -105.1 \\
Deterministic annealing &  -105.8 &  -97.3 & -101.4 & -94.3 \\
\gls{PVI}, Entropy constraint &-106.0 & -98.3 & \bfseries -99.9 & \bfseries -93.7\\
\gls{PVI}, Mean/variance constraint & -105.9 &-97.5 & -100.6 & -93.9 \\
\bottomrule
\end{tabular}
\vspace{1ex}
\caption{\textbf{\Acrlong{PVI} improves over \gls{VI} in a three-layer sigmoid belief network.}
The model has three layers of $200$~latent variables. We report the \glsreset{ELBO}\gls{ELBO} and marginal likelihood on the MNIST~\citep{pmlr-v15-larochelle11a}
validation set. \gls{PVI} and deterministic annealing~\citep{Katihara:2008} perform similarly for bad initialization; \gls{PVI} improves over deterministic annealing with good initialization.}

\label{table:sbn_3_layer}
\end{table}

%% file: tab_dlgm.tex

\begin{table}[tb]
\centering
\begin{tabular}{lSS}
\toprule
Inference method & \multicolumn{1}{c}{\gls{ELBO}} & \multicolumn{1}{c}{Marginal likelihood}
\\
\midrule
Variational inference & -101.0 & -94.2 \\
\gls{PVI}, Orthogonal constraint &\bfseries -100.3 &\bfseries -93.9 \\
\bottomrule
\end{tabular}
\vspace{1ex}
\caption{\textbf{\Acrlong{PVI} with an orthogonal proximity statistic makes optimization easier in a variational autoencoder model~\citep{kingma2014autoencoding,rezende2014stochastic}.} We report the held-out \glsreset{ELBO}\gls{ELBO} and estimates of the marginal likelihood on the binarized MNIST~\citep{pmlr-v15-larochelle11a} validation set.}
\label{table:dlgm2}
\end{table}

%% file: sec_discussion.tex
\section{Discussion}
We presented \acrlong{PVI}, a flexible method designed to avoid bad
local optima and to be robust to poor initializations.
We showed that \acrlong{VI} gets trapped
in bad local optima and cannot recover from poorly initialized parameters. The choice of proximity statistic $f$ and distance $d$ enables the design of a variety of
constraints. As examples of statistics, we gave the entropy, \gls{KL}
divergence, orthogonal proximity statistic, and the mean and variance of the approximate posterior. We evaluated our
method in three models to demonstrate that it is easy to
implement, readily extendable, and leads to beneficial statistical
properties of variational inference algorithms.

\paragraph{Future work.} Simplifying optimization is necessary for truly black-box variational inference.
An adaptive magnitude decay
based on the value of the constraint should improve results (this could even be
done per-parameter). New proximity constraints are also easy to design and test.
For example, the variance of the gradients of the variational parameters is a valid proximity statistic---which can be used to avoid variational approximations that have high variance gradients.
Another set of
interesting proximity statistics are empirical statistics of the variational
distribution, such as the mean, for when analytic forms are unavailable.
We also
leave the design and study of constraints that admit coordinate updates to
future work.

\subsubsection*{Acknowledgments}
The experiments presented in this article were performed on computational resources supported by the Princeton Institute for Computational Science and Engineering (PICSciE), the Office of Information Technology's High Performance Computing Center and Visualization Laboratory at Princeton University, Colin Raffel, and Dawen Liang. We are very grateful for their support. We also thank Emtiyaz Khan and Ben Poole for helpful discussions, and Josko Plazonic for computation support.

%% file: proximity_vi.bbl
\begin{thebibliography}{}

\bibitem[Bishop, 2006]{Bishop:2006}
Bishop, C.~M. (2006).
\newblock {\em Pattern Recognition and Machine Learning}.
\newblock Springer New York.

\bibitem[Bowman et~al., 2016]{DBLP:conf/conll/BowmanVVDJB16}
Bowman, S.~R., Vilnis, L., Vinyals, O., Dai, A.~M., J{\'{o}}zefowicz, R., and
  Bengio, S. (2016).
\newblock Generating sentences from a continuous space.
\newblock In {\em Conference on Computational Natural Language Learning}.

\bibitem[Boyd and Vandenberghe, 2004]{boyd2004convex}
Boyd, S. and Vandenberghe, L. (2004).
\newblock {\em Convex Optimization}.
\newblock {Cambridge University Press}.

\bibitem[Burda et~al., 2015]{Burda2016}
Burda, Y., Grosse, R., and Salakhutdinov, R. (2015).
\newblock {Importance weighted autoencoders}.
\newblock {\em International Conference on Learning Representations}.

\bibitem[Glorot and Bengio, 2010]{DBLP:journals/jmlr/GlorotB10}
Glorot, X. and Bengio, Y. (2010).
\newblock Understanding the difficulty of training deep feedforward neural
  networks.
\newblock In {\em Artificial Intelligence and Statistics}, pages 249--256.

\bibitem[{Higgins} et~al., 2016]{2016arXiv160605579H}
{Higgins}, I., {Matthey}, L., {Glorot}, X., {Pal}, A., {Uria}, B., {Blundell},
  C., {Mohamed}, S., and {Lerchner}, A. (2016).
\newblock {Early Visual Concept Learning with Unsupervised Deep Learning}.
\newblock {\em ArXiv:1606.05579}.

\bibitem[Hoffman et~al., 2013]{hoffman2013stochastic}
Hoffman, M.~D., Blei, D.~M., Wang, C., and Paisley, J. (2013).
\newblock Stochastic variational inference.
\newblock {\em Journal of Machine Learning Research}, 14:1303--1347.

\bibitem[Jordan et~al., 1999]{jordan1999introduction}
Jordan, M.~I., Ghahramani, Z., Jaakkola, T.~S., and Saul, L.~K. (1999).
\newblock An introduction to variational methods for graphical models.
\newblock {\em Machine Learning}, 37(2):183--233.

\bibitem[Katahira et~al., 2008]{Katihara:2008}
Katahira, K., Watanabe, K., and Okada, M. (2008).
\newblock Deterministic annealing variant of variational bayes method.
\newblock {\em Journal of Physics: Conference Series}, 95(1):012015.

\bibitem[Khan et~al., 2016]{Khan:2016:FSV:3020948.3020982}
Khan, M.~E., Babanezhad, R., Lin, W., Schmidt, M., and Sugiyama, M. (2016).
\newblock Faster stochastic variational inference using proximal-gradient
  methods with general divergence functions.
\newblock In {\em Uncertainty in Artificial Intelligence}, pages 319--328.

\bibitem[Khan et~al., 2015]{khan2015kullback}
Khan, M.~E., Baqu{\'e}, P., Fleuret, F., and Fua, P. (2015).
\newblock Kullback-leibler proximal variational inference.
\newblock In {\em Advances in Neural Information Processing Systems}, pages
  3384--3392.

\bibitem[Kingma and Ba, 2015]{DBLP:journals/corr/KingmaB14}
Kingma, D.~P. and Ba, J. (2015).
\newblock Adam: {A} method for stochastic optimization.
\newblock {\em International Conference on Learning Representations}.

\bibitem[Kingma and Welling, 2014]{kingma2014autoencoding}
Kingma, D.~P. and Welling, M. (2014).
\newblock Auto-encoding variational {B}ayes.
\newblock In {\em International Conference on Learning Representations}.

\bibitem[Kucukelbir et~al., 2015]{Kucukelbir:2015}
Kucukelbir, A., Ranganath, R., Gelman, A., and Blei, D.~M. (2015).
\newblock {Automatic Variational Inference in Stan}.
\newblock In {\em Neural Information Processing Systems}.

\bibitem[Larochelle and Murray, 2011]{pmlr-v15-larochelle11a}
Larochelle, H. and Murray, I. (2011).
\newblock The neural autoregressive distribution estimator.
\newblock In {\em Artificial Intelligence and Statistics}, volume~15, pages
  29--37.

\bibitem[MacKay, 2003]{MacKay:2003}
MacKay, D. (2003).
\newblock {\em Information Theory, Inference, and Learning Algorithms}.
\newblock Cambridge University Press.

\bibitem[Mansinghka et~al., 2014]{DBLP:journals/corr/MansinghkaSP14}
Mansinghka, V., Selsam, D., and Perov, Y.~N. (2014).
\newblock Venture: a higher-order probabilistic programming platform with
  programmable inference.
\newblock {\em arXiv:1404.0099}.

\bibitem[Mnih and Rezende, 2016]{Mnih:2016:VIM:3045390.3045621}
Mnih, A. and Rezende, D.~J. (2016).
\newblock Variational inference for monte carlo objectives.
\newblock In {\em International Conference on Machine Learning}, pages
  2188--2196.

\bibitem[Nakajima et~al., 2013]{nakajima2013}
Nakajima, S., Sugiyama, M., Babacan, S.~D., and Tomioka, R. (2013).
\newblock Global analytic solution of fully-observed variational bayesian
  matrix factorization.
\newblock {\em Journal of Machine Learning Research}.

\bibitem[Neal, 1992]{neal1992connectionist}
Neal, R.~M. (1992).
\newblock Connectionist learning of belief networks.
\newblock {\em Artificial intelligence}, 56(1):71--113.

\bibitem[Ranganath et~al., 2014]{ranganath2014black}
Ranganath, R., Gerrish, S., and Blei, D.~M. (2014).
\newblock Black box variational inference.
\newblock In {\em Artificial Intelligence and Statistics}.

\bibitem[Ranganath et~al., 2015]{ranganath2015deep}
Ranganath, R., Tang, L., Charlin, L., and Blei, D.~M. (2015).
\newblock Deep exponential families.
\newblock In {\em Artificial Intelligence and Statistics}.

\bibitem[Rezende and Mohamed, 2015]{Rezende:2015}
Rezende, D.~J. and Mohamed, S. (2015).
\newblock Variational inference with normalizing flows.
\newblock In {\em International Conference on Machine Learning}.

\bibitem[Rezende et~al., 2014]{rezende2014stochastic}
Rezende, D.~J., Mohamed, S., and Wierstra, D. (2014).
\newblock Stochastic backpropagation and approximate inference in deep
  generative models.
\newblock In {\em International Conference on Machine Learning}.

\bibitem[Saxe et~al., 2013]{saxe2013exact}
Saxe, A.~M., McClelland, J.~L., and Ganguli, S. (2013).
\newblock Exact solutions to the nonlinear dynamics of learning in deep linear
  neural networks.
\newblock {\em arXiv preprint arXiv:1312.6120}.

\bibitem[Spall, 2003]{Spall:2003}
Spall, J. (2003).
\newblock {\em Introduction to Stochastic Search and Optimization: Estimation,
  Simulation, and Control}.
\newblock Wiley.

\bibitem[Theis and Hoffman, 2015]{Theis2015}
Theis, L. and Hoffman, M.~D. (2015).
\newblock {A trust-region method for stochastic variational inference with
  applications to streaming data}.
\newblock {\em Journal of Machine Learning Research}.

\bibitem[Tran et~al., 2016]{tran2016edward}
Tran, D., Kucukelbir, A., Dieng, A.~B., Rudolph, M., Liang, D., and Blei, D.~M.
  (2016).
\newblock {Edward: A library for probabilistic modeling, inference, and
  criticism}.
\newblock {\em arXiv:1610.09787}.

\bibitem[Wainwright and Jordan, 2008]{wainwright2008graphical}
Wainwright, M.~J. and Jordan, M.~I. (2008).
\newblock Graphical models, exponential families, and variational inference.
\newblock {\em Foundations and Trends in Machine Learning}, 1(1-2):1--305.

\end{thebibliography}
